\documentclass[12pt]{article}
\usepackage{graphicx}

\usepackage{amsmath}

\usepackage{cite} 

\usepackage{subcaption}
\usepackage{array}
\usepackage{graphicx}

\usepackage{tikz}
\usetikzlibrary{shapes, arrows, positioning, fit, backgrounds}
\usetikzlibrary{intersections}
\usetikzlibrary{decorations, decorations.markings}
\usetikzlibrary{patterns,snakes, calc}

\tikzstyle{ground}=[preaction={fill, top color=black!10, bottom color=black!5, shading angle=25},
                    fill, pattern=north west lines,draw=none,minimum width=0.3,minimum height=0.6]
\tikzstyle{mass}=[line width=0.6, rounded corners=1,
                  top color=gray!40!black!20,bottom color=gray!40!black!10,shading angle=20]

\tikzstyle{spring}=[line width=0.8, snake=coil,segment amplitude=5,segment length=5,line cap=round]

\tikzstyle{damper}=[thick,decoration={markings,  
   mark connection node=dmp,
   mark=at position 0.5 with 
   {
     \node (dmp) [thick,inner sep=0pt,transform shape,rotate=-90,minimum
 width=15pt,minimum height=3pt,draw=none] {};
     \draw [thick] ($(dmp.north east)+(2pt,0)$) -- (dmp.south east) -- (dmp.south
 west) -- ($(dmp.north west)+(2pt,0)$);
     \draw [thick] ($(dmp.north)+(0,-5pt)$) -- ($(dmp.north)+(0,5pt)$);
   }
 }, decorate]

\tikzstyle{latent} = [circle, draw, fill=white, minimum height=1.5em]
\tikzstyle{input} = [circle, draw, fill=gray!25, minimum height=1.5em]
\tikzstyle{line} = [draw, -latex']

\usepackage{authblk}

\begin{document}

	\title{\bf Multi-task Equation Discovery}
        \author[1]{S.C.\ Bee}
		\author[1]{N.\ Dervilis}
 	      \author[1]{K.\ Worden}
        \author[2]{L.A.\ Bull}
	   \affil[1]{Dynamics Research Group, Department of Mechanical Engineering, University of Sheffield, Mappin Street, Sheffield S1 3JD, UK}
	   \affil[2]{School of Mathematics and Statistics, University of Glasgow, Glasgow, G12 8QQ, UK}

	\date{}
	\maketitle
	\thispagestyle{empty}

\subsection*{Abstract}
Equation discovery provides a grey-box approach to system identification by uncovering governing dynamics directly from observed data. However, a persistent challenge lies in ensuring that identified models generalise across operating conditions rather than over-fitting to specific datasets. This work investigates this issue by applying a Bayesian relevance vector machine (RVM) within a multi-task learning (MTL) framework for simultaneous parameter identification across multiple datasets. In this formulation, responses from the same structure under different excitation levels are treated as related tasks that share model parameters but retain task-specific noise characteristics. A simulated single degree-of-freedom oscillator with linear and cubic stiffness provided the case study, with datasets generated under three excitation regimes. Standard single-task RVM models were able to reproduce system responses but often failed to recover the true governing terms when excitations insufficiently stimulated non-linear dynamics. By contrast, the MTL-RVM combined information across tasks, improving parameter recovery for weakly and moderately excited datasets, while maintaining strong performance under high excitation. These findings demonstrate that multi-task Bayesian inference can mitigate over-fitting and promote generalisation in equation discovery. The approach is particularly relevant to structural health monitoring, where varying load conditions reveal complementary aspects of system physics.

\quad

\noindent\textbf{Keywords:} Sparse Regression, Equation Discovery, Multi-Task Learning, System Identification.

\section{Introduction}
Equation discovery has been applied to the field of system identification to identify equations of motion that capture the behaviour of measured systems, along with the parameters associated with each of their terms \cite{Brunton2016}. In practice, however, any selected model is only an approximation of reality, and uncertainty is inevitably present in the predictive equations \cite{Beck2010}. To address this, Bayesian approaches have been widely explored for equation discovery, which quantify parameter uncertainty \cite{WORDEN2012, FUENTES2019, FUENTES2021, Beck2010, ZHU2022}. 

A potential challenge in equation discovery is the risk of over-fitting, where the discovered model fits the training data closely but fails to represent the underlying physical system. The antithesis of over-fitting is generalisation, which can be encouraged via several strategies. One way to aid generalisation is via \emph{multi-task learning} (MTL), which learns multiple tasks simultaneously and exploits shared information to capture underlying commonalities. Caruana outlines several domains of MTL \cite{Caruana1997}, some of which have since been translated into the context of structural health monitoring (SHM) \cite{Bee2023}. 

In this paper, MTL is applied by treating datasets from the same structure under different excitation levels as separate, but related, tasks. By sharing parameters across tasks, the model exploits common features while still accounting for differences introduced by varying excitations. This design ensures that the model does not over-fit to one excitation condition (one dataset) and instead learns a representation of the system that is valid across multiple responses. 

This approach is valuable in SHM, where different loading conditions often highlight different aspects of the same underlying physics. For example, low-level excitations may reveal the system’s linear response, while higher excitations expose nonlinear effects. By training across multiple excitation scenarios simultaneously, the model is able to combine these complementary insights into a single, coherent description of the structure.



\section{System Identification with Bayesian Inference}
System identification characterises the dynamics of a system from observations of the evolution of states through time. The current work follows the approach proposed by \cite{Brunton2016} and treats equation discovery as sparse linear regression, known as SINDy.

For a multi-degree-of-freedom system, the system states are typically displacement and velocity, 
\begin{equation}
\boldsymbol{x} = 
\begin{bmatrix} 
x_1 \\ 
x_2 
\end{bmatrix} =
\begin{bmatrix} 
y \\ 
\dot{y} 
\end{bmatrix}
\end{equation}

\noindent where $y$ is displacement, $\dot{y}$ is velocity and $\quad\dot{}\quad$ denotes the first order differentiation with respect to time.

Within the framework of equation discovery, the above formulation is combined with a predefined dictionary, or design matrix, of $M$ candidate terms, 
\begin{equation}
\mathbf{D} = [d_1(\boldsymbol{X}), d_2(\boldsymbol{X}), ..., d_M(\boldsymbol{X})]
\label{eq:dictionary}
\end{equation}

\noindent where $\boldsymbol{X} = [\boldsymbol{x_1}, \boldsymbol{x_2}, ..., \boldsymbol{x_N}]$ and $\mathbf{D} \in R^ {N \times M}$. This corresponds to $N$ data-points and $M$ candidate components (features). The design matrix is intentionally over-defined, incorporating a wide range of potential components that could contribute to the equation of motion of the system.

By utilising shrinkage \cite{Tipping2001}, the algorithm identifies only the base elements that are most relevant, effectively revealing the governing dynamics of the system.

\subsection{RVM Formulation of SINDy}
\begin{figure}[ht]
  \centering

\begin{tikzpicture}

\node[latent] (alpha) {};
\node[latent, below = 10mm of alpha] (w) {};
\node[input, below=25mm of w] (y) {};

\node[input, above left=10mm and 18mm of y] (x) {};
\node[input, left = 30mm of y] (F) {};

\node[latent, above right = 2mm and 18mm of alpha] (a){};
\node[latent, below right = 2mm and 18mm of alpha] (b){};

\node[latent, right=27mm of y] (s) {};
\node[latent, above = 25mm of s] (lambda) {};

\node[left = 0.5mm of alpha.west] (lalpha) {$\alpha_m$};
\node[above left=0.5mm and 0.5mm of w.north west] (lw) {$w_m$};
\node[above right=0.5mm and 0.5mm of w.north east] (xw) {};
\node[above right=0.5mm and 0.5mm of y.north east] (ly) {$y_{n}$};

\node[above= 0.3mm of x.north] (lx) {$d_{m}(\boldsymbol{x_n})$};
\node[above right=0.5mm and 0.5mm of F.north east] (lf) {$F_n$};
\node[below left=4mm and 0.5mm of F.north east] (xf) {};

\node[right=0.5mm  of a.east] (la) {$a$};
\node[right=0.5mm  of b.east] (lb) {$b$};

\node[above right=0.5mm and 0.5mm of s.north east] {$\sigma$};
\node[above right=0.5mm and 0.5mm of lambda.north east] {$\lambda$};

\path [line] (alpha) -> (w);
\path [line] (w) -> (y);

\path [line] (x) -> (y);
\path [line] (F) -> (y);

\path [line] (a) -> (alpha);
\path [line] (b) -> (alpha);

\path [line] (s) -> (y);
\path [line] (lambda) -> (s);

\begin{scope}[on background layer]
    \node[fill, rounded corners=1ex, line = yellow, fill= white, fit=(xf) (ly) (y) (lx) (F), inner xsep=2mm, inner ysep = 2mm, label={[anchor=south east, xshift= -1mm, yshift=1mm]south east:$N$}]{};

    \node[rounded corners=1ex, line = black, fit=(lalpha) (lx) (xw) (x) (alpha), inner xsep=2mm, inner ysep = 2mm, label={[anchor=south east, xshift= -1mm, yshift=1mm]south east:$M$}]{};

    \end{scope}

\end{tikzpicture}
  
  \caption{A graphical representation of the RVM. Grey nodes correspond to observed variables and white nodes represent latent variables. The $M$ plate represents the number of candidate features and the $N$ plate represents the number of data points.}
  \label{fig:STLGRAPH}
  \end{figure}
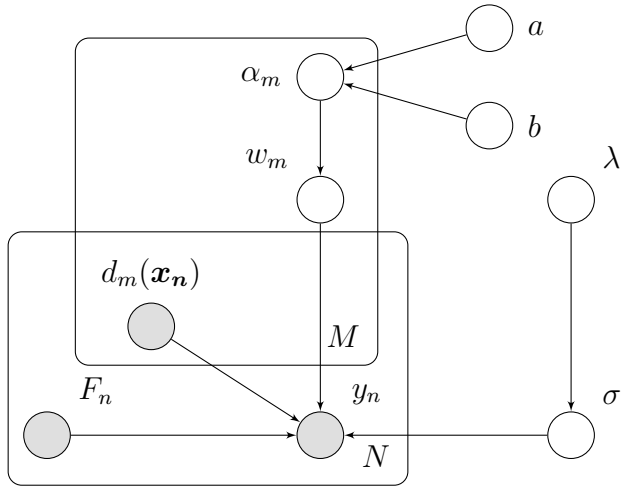

The relevance vector machine (RVM), devised by Tipping \cite{Tipping2001}, is a Bayesian algorithm which has a similar functional form to the deterministic support vector machine \cite{Kong2019}. RVMs have been used within various applications to analyse systems, and the approach is well-suited to equation discovery problems. The formulation used in this research is demonstrated graphically in Figure \ref{fig:STLGRAPH}.

In the context of identifying forced (non-autonomous) duynamical systems, a function $\mathbf{F}$ is present, which represents the input force acting on the structure. Arguably, this could be embedded within the design matrix, defined in equation (\ref{eq:dictionary}); however, in this work it is treated as a separate observed variable,  

 \begin{equation}
     \boldsymbol{y} = \mathbf{D} \mathbf{w} + \mathbf{F} +\boldsymbol{\epsilon}
     \label{eq:target}
 \end{equation}
where $\mathbf{w} \in R^M$ is the vector representing the weights associated with each candidate basis function and $\boldsymbol{\epsilon}$ is the noise that distorts the signal. 

Each weight is assumed to follow a zero-mean Gaussian prior, $w_m \sim \mathcal{N}(0,\alpha)$ where the variance is governed by an inverse gamma distribution, $\alpha_m^2 \sim \Gamma^{-1}(a, b)$. %

This hierarchical prior structure introduces hyper-parameters $a$ and $b$, which control the distribution of weight variances and encourage sparsity.

The signal noise is assumed to be Gaussian, $\epsilon \sim \mathcal{N}(0,\sigma^{2})$, with an exponential prior placed on the variance of the noise, $\sigma^{2} \sim \operatorname{Exp}(\lambda)$, where $\lambda$ is a hyper-parameter. This prior has a preference for data with generally low levels of noise. 

\subsection{RVM Extension to Multiple Tasks}
\begin{figure}[ht]
  \centering

\begin{tikzpicture}

\node[latent] (alpha) {};
\node[latent, below = 10mm of alpha] (w) {};
\node[input, below=25mm of w] (y) {};

\node[input, above left=10mm and 18mm of y] (x) {};
\node[input, left = 30mm of y] (F) {};

\node[latent, above right = 2mm and 18mm of alpha] (a){};
\node[latent, below right = 2mm and 18mm of alpha] (b){};

\node[latent, right=27mm of y] (s) {};
\node[latent, above = 25mm of s] (lambda) {};

\node[left = 0.5mm of alpha.west] (lalpha) {$\alpha_m$};
\node[above left=0.5mm and 0.5mm of w.north west] (lw) {$w_m$};
\node[above right=0.5mm and 0.5mm of w.north east] (xw) {};
\node[above right=0.5mm and 0.5mm of y.north east] (ly) {$y_{n,l}$};

\node[above= 0.3mm of x.north] (lx) {$d_{m}(\boldsymbol{X}_{n,l})$};
\node[above right=0.5mm and 0.5mm of F.north east] (lf) {$F_{n,l}$};
\node[below left=4mm and 0.5mm of F.north east] (xf) {};

\node[right=0.5mm  of a.east] (la) {$a$};
\node[right=0.5mm  of b.east] (lb) {$b$};

\node[above right=0.5mm and 0.5mm of s.north east] {$\sigma_l$};
\node[above right=0.5mm and 0.5mm of lambda.north east] {$\lambda$};

\path [line] (alpha) -> (w);
\path [line] (w) -> (y);

\path [line] (x) -> (y);
\path [line] (F) -> (y);

\path [line] (a) -> (alpha);
\path [line] (b) -> (alpha);

\path [line] (s) -> (y);
\path [line] (lambda) -> (s);

\begin{scope}[on background layer]
    \node[fill, rounded corners=1ex, line = yellow, fill= white, fit=(y) (ly) (lx) (xf) (s), inner xsep=6mm, inner ysep = 4mm, label={[anchor=south east, xshift= -1mm, yshift=1mm]south east:$L$}]{};
    
    \node[fill, rounded corners=1ex, line = yellow, fill= white, fit=(xf) (ly) (y) (lx) (F), inner xsep=2mm, inner ysep = 2mm, label={[anchor=south east, xshift= -1mm, yshift=1mm]south east:{$N_l$}}]{};

    \node[rounded corners=1ex, line = black, fit=(lalpha) (lx) (xw) (x) (alpha), inner xsep=2mm, inner ysep = 2mm, label={[anchor=south east, xshift= -1mm, yshift=1mm]south east:$M$}]{};
    
    \end{scope}

\end{tikzpicture}
  
  \caption{A graphical representation of a multi-task RVM. Grey nodes correspond to observed variables and white nodes represent latent variables. The $L$ plate represents the number of tasks, the $M$ plate represents the number of candidate features and the $N$ plate represents the number of data points}
  \label{fig:GRAPH}
  \end{figure}
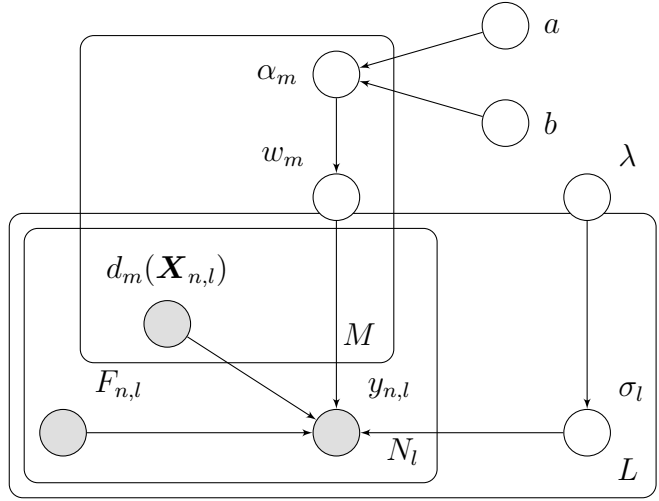

When analysing $L$ tasks, the RVM can be adapted as shown in Figure \ref{fig:GRAPH}. 

Figure \ref{fig:GRAPH} shows that the weight vector, $\mathbf{w}$, is shared between \emph{all} tasks, while the noise variance is task-specific. A shared weight distribution reflects the assumption that measurements across tasks have common underlying physical properties. In contrast, the noise is modelled separately for each task as it typically arises from environmental conditions or measurement settings that vary between tasks.

\section{Case Study}
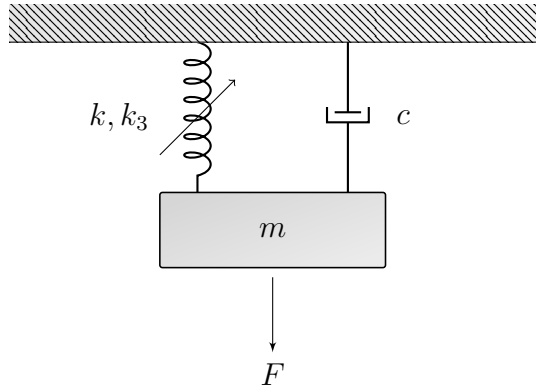
\begin{figure}[ht]
  \centering
  
  \begin{tikzpicture}
    \def\H{0.5} 
    \def\W{7.0}  
    \def\h{1}  
    \def\w{3}  
    \def\y{2}  
  
    \draw[spring,segment length=7] (-1,0) -- (-1,-\y);
    \draw[ground] (-\W/2,0) rectangle++ (\W,\H);
    \draw (-\W/2,0) --++ (\W,0);
    \draw[mass] (-\w/2,-\y) rectangle++ (\w,-\h) node[midway] (m) {$m$};
  
    \draw[damper] (1,-\y) -- (1,0);
  
    \node[anchor=west] at (1.5, -\y/2) {$c$};
    
    \draw[->] (-1.5,-1.5) -- (-0.5,-0.5);
    \node[anchor=east] at (-1.5, -\y/2) {$k, k_3$};

    \node[below = 1mm of m.south](m2){};
    \node[below = 10mm of m2.south](F){$F$};
    \path [line] (m2) -> (F);

  \end{tikzpicture}
  
  \caption{SDOF oscillator with stiffness (linear, $k$ and cubic, $k_3$), mass $m$, damping coefficient $c$ and force acting on the system, $F$.}
  \label{fig:SDOF}
  \end{figure}

A simulated dataset was developed, based on a single degree-of-freedom (SDOF) system with nonlinear stiffness, shown by Figure \ref{fig:SDOF}. 

The equation of motion for this system is given by,
  \begin{equation}
    m\ddot{y} + c \dot{y} + k_1y + k_3y^3 = F
      \label{eq:motion}
  \end{equation}
  
\noindent where $m = 1 kg$ is mass, $c = 0.2Ns/m$ is the damping coefficient, $k_1=1 N/m$ is the linear stiffness, $k_3 = 1 N/m^3$ is the cubic stiffness, $F$ is the system forcing, and, $\quad\ddot{}\quad$ denotes the second order differentiation with respect to time. The values for the damping coefficient and stiffnesses were selected to demonstrate the algorithm and not to be representative of a real system. 

The system was integrated forward in time via a fourth-order Runge-Kutta method, with a sampling frequency of 100kHz. To reduce dataset size, the signal was downsampled by selecting every thousandth point, resulting in an effective sampling frequency of 100Hz over an interval of 10s. A Butterworth filter \cite{Butterworth1930} was applied with a low cut-off of 3Hz. 

To generate a suitable dataset, it was necessary to construct data under varying conditions. Since the objective is to use these datasets to gain insight into the structure, the underlying physics, represented by the right-hand side of equation (\ref{eq:motion}), must remain unchanged. Consequently, only the input forcing can be modified to produce different datasets. The forcing is modelled as random excitation in the form of noise, drawn from a uniform distribution whose amplitude can be scaled by a factor, $f$, such that $F \sim f \, \mathcal{U}(-0.5, 0.5)$. To create three distinct datasets, data were generated over a duration of 10s with $f = \{10^1, 10^2, 10^3\}$. Three datasets allow for a single task (ST) scenario with low, medium and high excitation, one at each of the values of $f$, and one multi-task (MT) scenario with all three datasets utilised in a joint inference for the identification of the shared underlying system. 

For each of the three datasets, the dictionary of candidate basis functions required for the RVM was defined as,

\begin{equation}
D(\mathbf{X}) = 
\begin{bmatrix}
\vert & \vert & \vert & \vert & \vert & \vert & \vert\\
\mathbf{y} & \mathbf{\dot y} & \mathbf{y}^2 & \mathbf{\dot y}^2 & \mathbf{y}^3 & \mathbf{\dot y}^3 & \mathbf{1} \\
\vert & \vert & \vert & \vert & \vert & \vert & \vert\\
\end{bmatrix}
\end{equation}

\section{Results and Discussion}

It is worth noting that the target vector, $\boldsymbol{y}$ (equation (\ref{eq:target})), for this problem is acceleration, $\ddot{y}$. Hence, Equation (\ref{eq:motion}) will take the form, 

\begin{equation}
    m\ddot{y}  = F - c \dot{y} - k_1y - k_3y^3
\end{equation}

\noindent Therefore, the target value of $\boldsymbol{w}$ will be $[-k_1 \quad -c \quad 0 \quad 0 \quad -k_3 \quad 0 \quad 0]$. 

\begin{figure}[t]
    \centering
    \includegraphics[width=0.8\linewidth]{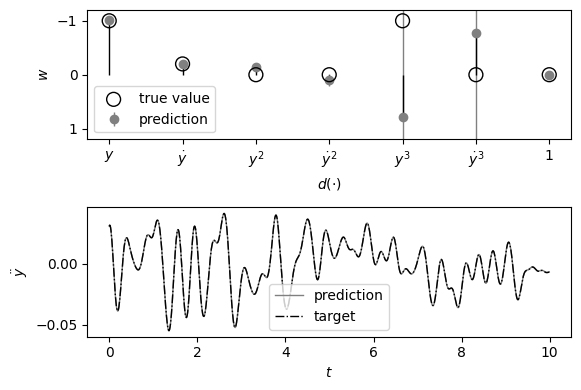}
    \caption{Results of the ST scenario with low excitation, $f=10^1$, using a standard RVM (as per Figure \ref{fig:STLGRAPH}); (\emph{Upper}) Weight matrix, $\boldsymbol{W}$, results; (\emph{Lower}) the predicted response.}
    \label{fig:STL1}
\end{figure}

Figure \ref{fig:STL1} shows the results for $f=10^1$. The lower plot shows that the RVM output tracks the target acceleration over the 10-second training dataset reasonably well. However, the estimated values of the elements in $\boldsymbol{w}$, shown in the upper plot, deviate notably from the true values.

\begin{figure}[ht]
    \centering
    \begin{subfigure}{\linewidth}
        \centering
        \includegraphics[width=0.8\linewidth]{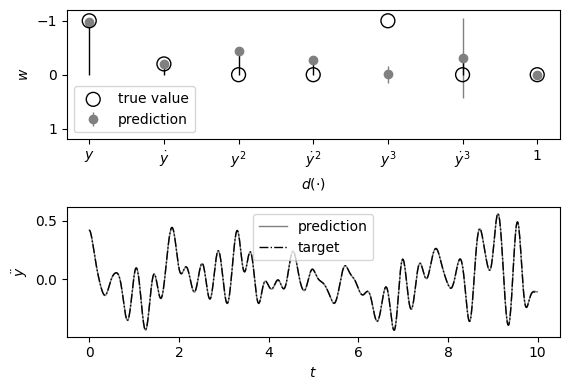}
        \caption{Results of the ST scenario with medium excitation $f=10^2$ using a standard RVM.}
        \label{fig:STL2}
    \end{subfigure}
    
    \vspace{1em} 
    
    \begin{subfigure}{\linewidth}
        \centering
        \includegraphics[width=0.8\linewidth]{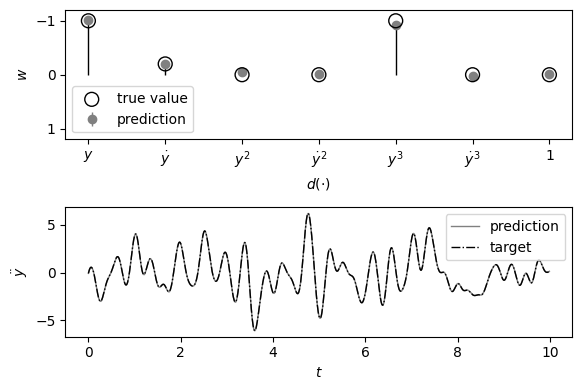}
        \caption{Results of the ST scenario with high excitation, $f=10^3$ using a standard RVM.}
        \label{fig:STL3}
    \end{subfigure}
    
    \caption{Results of the ST scenario with medium and high excitation using a standard RVM (as per Figure \ref{fig:STLGRAPH}). (\emph{Upper}) Weight matrix, $\boldsymbol{W}$, results; (\emph{Lower}) the predicted response.}
    \label{fig:STL2and3}
\end{figure}

\clearpage
\newpage

The mean of the displacement cubic term, $y^3$, is 0.79, much closer to one than the true value of minus one and the mean of the velocity cubic term, $\dot{y}^3$, is -0.77, closer to -1 than the true value of 0, this outcome is challenging to interpret in physical terms.

Figure \ref{fig:STL2} shows the results for $f=10^2$ and Figure \ref{fig:STL3} shows the results for $f=10^3$. As with the lower plot in Figure \ref{fig:STL1}, the corresponding plots in both Figure \ref{fig:STL2} and Figure \ref{fig:STL3} accurately tracks the target acceleration over the 10-second training dataset. 

However, for $f=10^2$ the estimated weight matrix again diverges from the \emph{True} values. In this case, $y^3$, is 0.01, much closer to zero than the true value of -1, this discrepancy can likely be attributed to the insufficient excitation of the system’s non-linearities. The terms $y^2$, $\dot{y}^2$, and $\dot{y}^3$ all take on values larger than $\dot{y}$ (which was correctly identified as $0.2$), again these outcomes are difficult to interpret in physical terms.

\begin{figure}[p]
    \centering
    \includegraphics[width=0.8\linewidth]{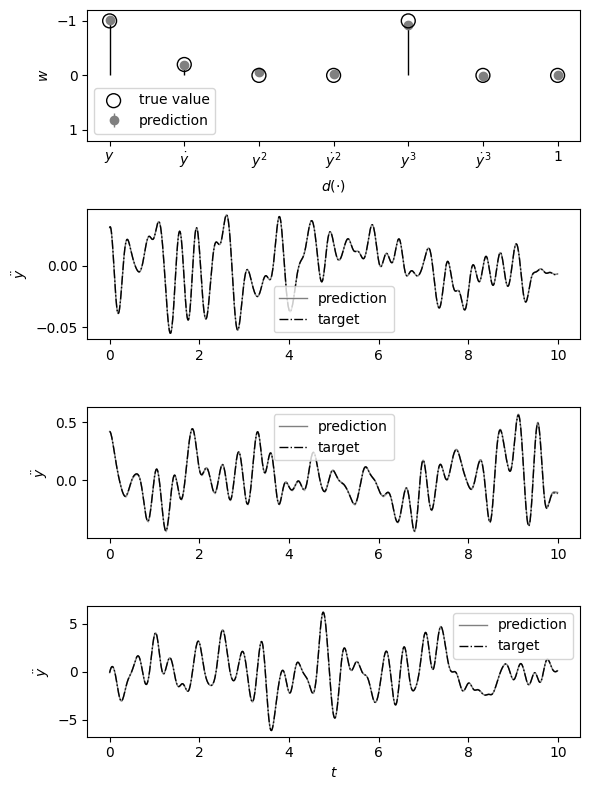}
    \caption{Results of the MT scenario with $f=[10^1 \quad 10^2 \quad 10^3]$ using a multi-task RVM (as per Figure \ref{fig:GRAPH}); (\emph{Upper}) Weight matrix, $\boldsymbol{W}$, results shared across all tasks; (\emph{Second}) the predicted response for $f=10^1$; (\emph{Third}) the predicted response for $f=10^2$; and, (\emph{Fourth}) the predicted response for $f=10^3$.}
    \label{fig:MTL}
\end{figure}

The weight matrix for $f=10^3$ is in good agreement with the \emph{True} values of $\boldsymbol{W}$, the value for $y^3$ is noticeably slightly lower than the \emph{True} value but is much closer than when $f=10^2$ and $f=10^3$. 

The results for the combination of all the data as three distinct tasks is shown in Figure \ref{fig:MTL}. The prediction of values over the 10s interval for the three tasks remains good and there is improvement in the prediction of $W$ for $f=10^1$ and $f=10^2$. The results of $f=10^3$ and the multi-task learner are similar indicating that there has been positive transfer from $f=10^3$ to the other two tasks, without negative transfer impacting the strongest performing task.

The parameter used to evaluate the predictive performance of each model is the normalised mean-square error metric: 

\begin{equation}
    NMSE =\frac{100}{N \sigma_z^2} \Sigma_{n=1}^N (z_n - \hat z_n)^2
\end{equation}
 \noindent $\sigma_z^2$ is the variance of the target ($z$), $\hat{\cdot}$ represents the predicted value, and, in this study $\boldsymbol{z} = \boldsymbol{\ddot{y}}$. 

\begin{figure}[t]
    \centering
    \includegraphics[width=\linewidth]{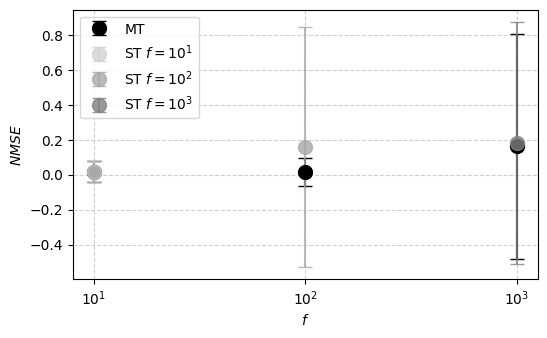}
    \caption{The distribution of NMSE for the MT and ST scenarios showing $3\sigma$}
    \label{fig:NMSE}
\end{figure}

Figure \ref{fig:NMSE} shows the NMSE results from the ST and MT scenarios for the three excitation levels. One hundred datasets were generated at each excitation level and the NMSE for each dataset was calculated. The graph shows the average and distribution of the NMSE for the different excitation levels and the different models. 

For low and high excitation case, the NMSE is similar between the ST and MT models. The benefit of the MT model is demonstrated in the medium excitation case as the average NMSE is lower, but also, the variance is lower. With medium excitation, the data does not contain information on the non-linear components of the structure, as these non-linear components have not been excited. Therefore, the ST model with medium excitation has been unable to capture this (aforementioned) and instead, has fit a model based on the incomplete data. Hence, the ST model's weight matrix does not match well with actual parameters. In contrast, the MT model \emph{does} have a dataset which includes both linear and non-linear excitation, the parameters are much closer to the \emph{True} values. 

\section{Conclusion}
This provides a strong example of how multi-task learning can enhance equation discovery by promoting generalisation across tasks. When working with a single task, it is often difficult to assess whether the identified model truly generalises. For instance, the solutions obtained for $f=10^1$ and $f=10^2$ differ considerably between the two single-task models. By contrast, the multi-task RVM yields a solution that performs well across all three datasets. This consistency suggests that the model has captured a representation that better reflects the underlying physics of the system. When considering the NMSE of the models, the MT model outperforms the ST models with a lower standard deviation in the medium and high excitation cases. 

\section*{Acknowledgements}
The authors wish to gratefully acknowledge support for this work through grants from the Engineering and Physical Sciences Research Council (EPSRC), UK, and Natural Environment Research Council, UK via grant number, EP/S023763/1. For the purpose of open access, the author has applied a Creative Commons Attribution (CC BY) licence to any Author Accepted Manuscript version arising.


\bibliography{Bee}

\end{document}